\documentclass[journal]{IEEEtran}

\usepackage{caption}
\usepackage{graphicx}
\usepackage{amsmath,amssymb} % define this before the line numbering.
\usepackage{color}
\usepackage{multirow}
\usepackage{mathrsfs}
\usepackage{subfigure}
\usepackage{array}
\usepackage{threeparttable}
\usepackage{float}
\usepackage{cite}
\usepackage[colorlinks]{hyperref}
\usepackage{algorithm}
\usepackage{algpseudocode}
\usepackage{eqparbox}
\usepackage{bm}

\ifCLASSINFOpdf

\else

\fi

\begin{document}

	\title{Progressive Graph Convolution Network for EEG Emotion Recognition}
	\author{Yijin Zhou, 
		Fu Li,~\IEEEmembership{Member,~IEEE,}
		Yang Li$^*$,~\IEEEmembership{Member,~IEEE,}
		Youshuo Ji, 		
		Guangming Shi,~\IEEEmembership{Fellow,~IEEE,} 
		Wenming Zheng,~\IEEEmembership{Senior Member,~IEEE,}
		Lijian Zhang, 	
		Yuanfang Chen, 
		Rui Cheng
	\thanks{Yijin Zhou, Fu Li, Yang Li, Youshuo Ji, Guangming Shi, and Rui Cheng are with the Key Laboratory of Intelligent Perception and Image Understanding of Ministry of Education, the School of Artificial Intelligence, Xidian University, Xi’an, 710071, China.\it{($^*$Corresponding author: Yang Li (E-mail: liy@xidian.edu.cn).)} \protect}% <-this % stops a space
	\thanks{Wenming Zheng is with the Key Laboratory of Child Development and Learning Science (Ministry of Education), School of Biological Sciences and Medical Engineering,
	Southeast University, Nanjing, Jiangsu, 210096, China.\protect }% <-this % stops a space
	\thanks{Lijian Zhang and Yuanfang Chen are with the Beijing Institute of Mechanical Equipment, Beijing, 100000, China. \protect }	
	}
	
	\maketitle
	\begin{abstract}
		Studies in the area of neuroscience have revealed the relationship between emotional patterns and brain functional regions, demonstrating that dynamic relationships between different brain regions are an essential factor affecting emotion recognition determined through electroencephalography (EEG). Moreover, in EEG emotion recognition, we can observe that clearer boundaries exist between coarse-grained emotions than those between fine-grained emotions, based on the same EEG data; this indicates the concurrence of large coarse- and small fine-grained emotion variations. Thus, the progressive classification process from coarse- to fine-grained categories may be helpful for EEG emotion recognition. Consequently, in this study, we propose a progressive graph convolution network (PGCN) for capturing this inherent characteristic in EEG emotional signals and progressively learning the discriminative EEG features. To fit different EEG patterns, we constructed a dual-graph module to characterize the intrinsic relationship between different EEG channels, containing the dynamic functional connections and static spatial proximity information of brain regions from neuroscience research. Moreover, motivated by the observation of the relationship between coarse- and fine-grained emotions, we adopt a dual-head module that enables the PGCN to progressively learn more discriminative EEG features, from coarse-grained (easy) to fine-grained categories (difficult), referring to the hierarchical characteristic of emotion. To verify the performance of our model, extensive experiments were conducted on two public datasets: SEED-IV and multi-modal physiological emotion database (MPED). The experiment results show that the PGCN achieves a state-of-art performance. Furthermore, we explored the effect of different frequency bands based on our model, and visualized the activated brain regions. The experiment results reveal the relationship between human emotion and high-frequency EEG signals, as well as the importance of the frontal and temporal lobes for emotion expression.
	\end{abstract}
	
	
	\IEEEpeerreviewmaketitle
	
	\section{Introduction}
		
		Emotions play an extremely important role in human\textendash human interaction activities, such as negotiation and work~\cite{2009Emotion, doi:10.1080/23808985.1994.11678894}. Accordingly, in a human\textendash machine interaction (HMI), we hope that machines can make adaptive changes according to the variation of human emotions to communicate with us, thus making emotion recognition an important topic in HMI~\cite{911197}. Human beings can easily realize the emotional states of others through their behavioral signals. Therefore, many studies on emotion recognition have focused on distinguishing emotional states according to behavioral signals, such as changes in voice signals~\cite{2003Emotion, 1202279} and facial expressions~\cite{wang2018intelligent}. Compared with these behavioral signals, physiological signals can also reflect human emotional changes, but are more reliable and difficult to camouflage~\cite{9373917}. In various physiological signals, electroencephalography (EEG) signals are of particular concern because of their non-invasiveness, portability, and affordability, and have received substantial attention recently~\cite{lawhern2018eegnet, soroush2017review, berti2019early}.
		
		\begin{figure}[t]
			\centering 
			\subfigure{\includegraphics[width=0.47\linewidth]{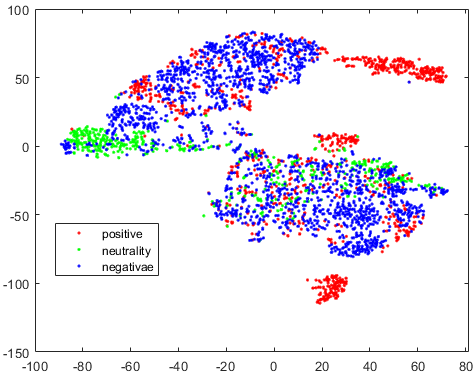}}
			\quad
			\subfigure{\includegraphics[width=0.47\linewidth]{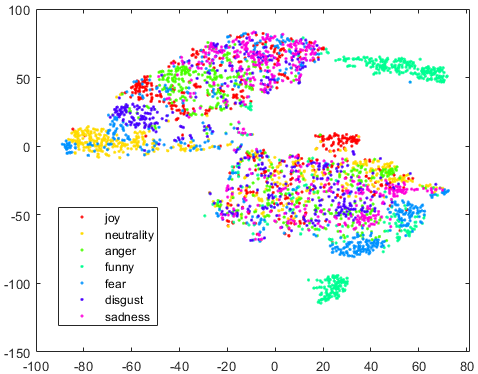}}
			\caption{\label{TSNE_hie}Distribution of raw EEG data on multi-modal physiological emotion database (MPED): (a) coarse-grained data containing three emotions and (b) fine-grained data containing seven emotions.}
		\end{figure}
		
		Two issues should be addressed for EEG emotion recognition. The first issue is how to utilize the inherent characteristics of emotions to extract more discriminative features. Several popular methods for EEG emotion recognition mainly utilize the general properties of EEG signals, rather than the characteristics of emotions in such signals. For example, Alhagry et al. proposed a long short-term memory (LSTM) network with a dense layer for extracting temporal relations and achieving the classification of EEG emotion recognition~\cite{alhagry2017emotion}. In addition, Jia et al. proposed SST-EmotionNet for EEG emotion recognition, which is a novel dense 3D spatial-spectral-temporal-based attention network~\cite{jia2020sst}. These methods have not sufficiently exploited the specific characteristics of emotions for EEG emotion recognition. Further, it would be more promising and fruitful if we could integrate the unique characteristics of emotion into deep learning methods to extract more emotion-related features. As an intrinsic characteristic of emotion, hierarchical characteristics are intuitively reflected in daily life. For example, when we communicate with others, it is easy to understand negative emotional states of other people (coarse-grained category); however, it is relatively difficult to further distinguish which negative emotion (fine-grained category), such as sadness, fear, or disgust, a person is experiencing. Some studies have also revealed this hierarchical characteristic for both coarse- and fine-grained categories. For example, Zhang et al. believe that positive emotions can be regarded as a class of emotions, including pride, contentment, amusement, love, and so on~\cite{2020CPED}. Belinda et al. believe these lower-level positive emotion categories are under the umbrella of happiness, sharing the same feeling of happiness together~\cite{doi:10.1080/02699931.2012.683852}. In addition, Philip et al. discovered that the Euclidean distances of disgust, fear, and anger are extremely similar, which shows the hierarchical characteristic of negative emotions from the other side~\cite{KRAGEL2016444}. To clearly demonstrate this hierarchical characteristic of emotion, we show the data distribution of coarse- and fine-grained emotions, based on the same EEG data, as indicated in Fig.~\ref{TSNE_hie}. Notably, the boundaries between coarse-grained emotions (Fig.~\ref{TSNE_hie} (a)) are clearer than those between fine-grained emotions (Fig.~\ref{TSNE_hie} (b)), which reflects the existence of the hierarchical characteristic of emotion. Therefore, this hierarchical characteristic is a type of emotion-specific information that can be exploited to obtain more discriminative information for EEG emotion recognition.
		
		The second issue for EEG emotion recognition is how to make use of the topological information of brain regions to model EEG emotional signals. Traditional methods are mainly based on convolutional neural networks (CNNs) and recurrent neural networks (RNNs), which model EEG signals as image- or speech-like data~\cite{8304360, 8275511}. However, owing to their characteristics, CNNs and RNNs are limited in that the model input must be grid data, ignoring the connection among brain regions. Recently, graph-based methods have also been proposed to handle EEG emotion recognition tasks, wherein the data are distributed within a non-Euclidean space~\cite{9091308, 8758787}. However, these methods only adopt a single or fixed graph structure to model the EEG signals, which is unsuitable for reflecting the internal relationships between brain regions and human emotions. Neuroimaging studies have shown that emotions have a unique relationship with the activity of distributed neural systems that span the cortical and subcortical regions~\cite{KRAGEL2016444}. Neuroscience studies have revealed the relationship between emotional patterns and the functional connections of the brain regions~\cite{KOBER2008998, KIM2011403}, and have indicated that the physical proximity of the brain in space is equally important~\cite{10.1093/cercor/bhi016}. Meanwhile, EEG signals vary among individuals, and the relationship between EEG electrodes is dynamic~\cite{9373917}. Therefore, it will be helpful to model EEG emotional signals according to the dynamic functional connections of the brain regions and prior knowledge in the area of neuroscience.
		
		To address the aforementioned issues, in this study, we propose a progressive graph convolution network (PGCN) to progressively learn the discriminative information among EEG emotion signals concerning EEG emotion recognition. To achieve the first goal, we need to solve the first issue, this is, how to utilize the hierarchical characteristic of emotion to extract more discriminative features. Specifically, the hierarchical characteristic of emotion contains information on the commonness and individuality between fine- and coarse-grained emotions. Motivated by this, we construct a dual-head fine- and coarse-grained module, which enables progressive emotion recognition from coarse-grained (easy) to fine-grained categories (difficult). The coarse-grained head extracts the common features of polarization emotions, this is, positive, negative, and neutral emotions, aiming to obtain coarse-grained emotion patterns. The fine-grained head captures particular features from the subtler emotions to encode fine-grained emotion patterns. The key to solving the second issue lies in encoding the functional connections of the brain regions corresponding to emotional patterns. In the procedure used to build the graph structure for modeling EEG signals, we construct a dual-graph module, comprising two brain view graphs based on the spatial proximity and functional connectivity of the brain. The functional connectivity-based dynamic graph originates entirely from the input instance and adaptively changes along with it, which regards the dependency of each pair of electrodes as directed. The spatial distance-based static graph considers the interaction relationship between adjacent brain regions at physical distances. By integrating the above two graphs, the PGCN can learn the complementary interactive information of all electrodes from a global perspective, which enhances the robustness and discrimination of the model.
		
		Our contributions are three-fold.
		
		$ \bullet $ We propose a dual-head model called a PGCN based on the hierarchical characteristic of emotion. The PGCN progressively learns discriminative EEG features from coarse- to fine-grained categories.
		
		$ \bullet $ We constructed a dual-graph module based on the functional connectivity and spatial proximity of the brain regions. The complementarity of the two brain view graphs provides rich spatial topology information for capturing the internal relationship between different EEG channels.
		
		$ \bullet $ We compared the PGCN with several advanced models on two public datasets, namely, SEED-IV and MPED. The results show that the proposed PGCN model achieves a state-of-art performance. The comparative experiment showed that human emotions are related more to high-frequency EEG signals.
		
		The rest of this paper is organized as follows. Section~\ref{Sec:Related works} summarizes related studies on EEG emotion recognition. Section~\ref{Sec: The proposed method} introduces the proposed PGCN in detail. Section~\ref{Sec: Experiment result} discusses the results of the extensive experiments conducted. Finally, Section~\ref{Sec: Conclusion} provides the concluding remarks.
		
	\section{Related Work}
	\label{Sec:Related works}
	
	\subsection{Existing Methods for EEG Emotion Recognition}
		
		In the past years, numerous studies have attempted to solve EEG emotion recognition using machine learning methods. For instance, Nie et al. used a linear dynamic system to smooth the features extracted from EEG data followed by a support vector machine (SVM) to classify positive and negative emotions. They demonstrated that emotion is closely related to different brain regions and frequency bands~\cite{5910636}. In addition, Lin et al. extracted five-band power spectrum features from EEG data collected during the music wake-up paradigm, and then, classified these features using a multi-layer perceptron (MLP) and an SVM to achieve classification~\cite{5458075}. Moreover, Gupta et al. comprehensively investigated a flexible analytic wavelet transform (FAWT) based feature extraction method for an emotional EEG, leading to the channel-specific nature of EEG signals~\cite{8546787}. Recently, with the outstanding performance of deep learning models for various tasks, such as target detection, target recognition, and speech recognition~\cite{Chen_2020_CVPR, Han_2020_CVPR, gulati2020conformer}, an increasing number of researchers have adopted deep learning methods in the field of EEG emotion recognition. Alhagry et al. proposed an LSTM network with a dense layer to extract temporal relations and achieve classification~\cite{alhagry2017emotion}. Xing et al. applied a stack autoencoder (SAE) and an LSTM network to build a linear hybrid model and an emotional time series model, respectively. These two models were then combined to classify EEG data from the DEAP dataset~\cite{10.3389/fnbot.2019.00037}. Chao et al. attempted to extract several EEG features using a multiband feature matrix (MFM), and a capsule network(CapsNet) was then employed for classification~\cite{s19092212}. In addition, Li et al. took advantage of the emotional brain asymmetries between the left and right hemispheres, and adopted a bi-hemisphere domain adversarial neural network (BiDANN) for EEG emotion recognition. They mapped the EEG data of the left and right hemispheres into discriminative feature spaces separately; employed a global and two local domain discriminators; and achieved high-quality performance on the SEED~\cite{8567966}. Based on the discrepancy in emotion expression between the left and right hemispheres of the human brain, Li et al. also proposed a novel bi-hemispheric discrepancy model (BiHDM) for EEG emotion recognition, and achieved excellent performance on three public EEG emotional datasets~\cite{9105104}. Nevertheless, although the aforementioned methods have achieved relatively acceptable results, how to model the relationship between EEG electrodes needs to be studied further.
	
	\subsection{GCN for EEG Emotion Recognition}
		
		Because different brain regions are in a non-Euclidean space, a graph becomes the most appropriate data structure for EEG signals. Hence, an increasing amount of recent research has focused on the provisioning of graph theory-based methods for EEG emotion recognition. Wang et al. developed a methodology called the broad dynamical graph learning system (BDGLS) for dynamically extracting features~\cite{8621147}. In different study, Wang et al. used a phase-locking value (PLV) based graph convolutional neural network (P-GCNN) model to acquire spatiotemporal characteristics and internal information~\cite{8758787}. Song et al. proposed a dynamical graph convolutional neural network (DGCNN) to recognize EEG emotion, which can dynamically learn the information transfer relationship between the nodes in a graph. A DGCNN was evaluated on the SEED and DREAMER dataset, and achieved good results~\cite{8320798}. Song et al. also introduced the instance-adaptive graph (IAG) method. The graph of the IAG was generated from the input samples to retain the spatial features of the EEG data. IAG successfully solved individual differences, and dynamically described the relationships among different EEG regions~\cite{2020Instance}. Yin et al. investigated a combination of a GCNN and an LSTM. The GCNN was adopted to extract graph domain features, while the LSTM was used to obtain temporal correlation~\cite{YIN2021106954}. The above studies demonstrate the importance of properly modeling EEG signals. Some researchers have noticed the significance of the dynamic functional connectivity of the brain regions. These aforementioned methods were compared with our model in the experimental study.
		
	\section{Proposed Method for Emotion Recognition}
	\label{Sec: The proposed method}
		
		\begin{figure*}[htb]
			\centering{\includegraphics[width=1\linewidth] {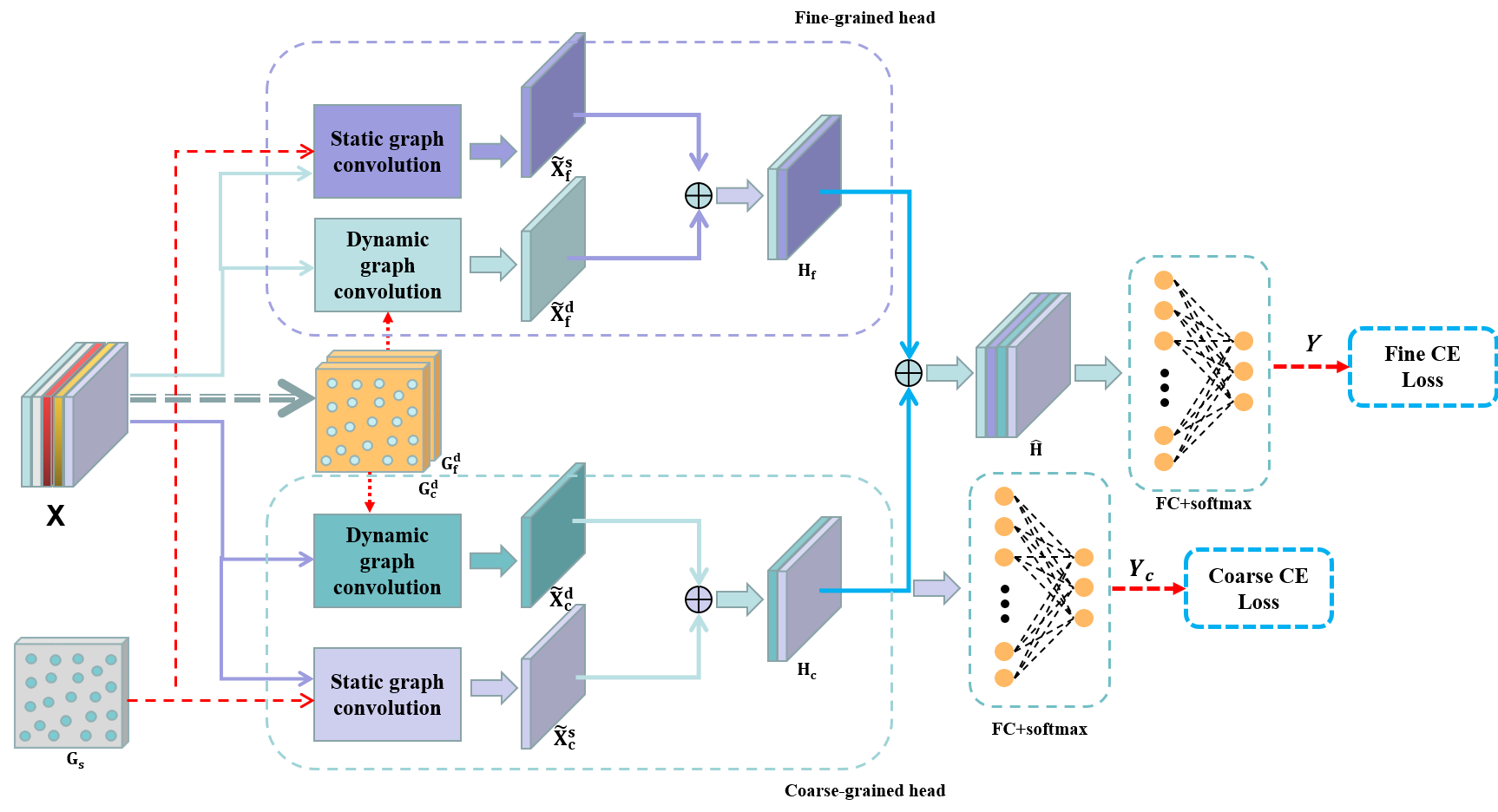}}
			\caption{\label{1} Framework of PGCN. The coarse-grained head is employed to extract the commonalities of emotional patterns, and the fine-grained head is employed to extract the subtle differences of the emotional patterns.}
		\end{figure*}
		
		To specify the proposed method clearly, we depict the PGCN framework in Fig.~\ref{1}. The network aims to capture the inherent characteristics of EEG emotional signals and learn the discriminative EEG features progressively. We adopted three steps to achieve this goal. The first step was to obtain a dynamic and static graph representation. Subsequently, the dual-head network was adopted to obtain deep emotion features, to capture the emotion patterns. Finally, we applied a dual-granular emotion classification progressively from coarse- to fine-grained categories. In the following section, we introduce the details of the proposed PGCN model.
		
		\subsection{Obtaining Dynamic and Static Graph Connections}
		
			To model the connections between different brain regions, we constructed two brain view graphs based on the spatial proximity and functional connectivity of the brain regions, to represent the dependency of each pair of nodes. Specifically, inspired by the attention mechanism~\cite{vaswani2017attention}, the PGCN learns a dynamic mapping relation to adaptively weigh the EEG channels according to the different instances, which constructs a dynamic graph that can select more useful electrodes in the brain regions. This graph is more powerful in representing the relationships among different EEG channels across different instances. Meanwhile, the physical proximity of the brain in space based on neuroscience research is an important reference that should be considered~\cite{10.1093/cercor/bhi016}. Inspired by this, we constructed a static graph that fully considers the spatial proximity, based on prior knowledge in the field of neuroscience.
			
			Specifically, we first decomposed the EEG signal into five frequency bands: $\delta$ band (1\textendash 5 Hz), $\theta$ band (4\textendash 8 Hz), $\alpha$ band (8\textendash 14 Hz), $\beta$ band (14\textendash 30 Hz) and $\gamma$ band (30\textendash 50 Hz). Subsequently, we extracted the energy feature from each frequency band to form the input EEG feature matrix $ {\rm \mathbf{X}} \in \mathbb{R} ^ {n\times d} $, where $n$ is the number of EEG channels, and $d$ is the number of frequency bands. We constructed a directed graph, represented as $ \mathcal{G} = (\mathcal{V}, \mathcal{E}) $, using the standard symbol representation in graph theory, where $ \mathcal{V} = {\{v_i\}}_{i=1}^{n} $ represents the set of $ n $ nodes in the graph, and $ (v_i,v_j) \in \mathcal{E} $ denotes the edge connected by nodes $ v_i $ and $ v_j $ in the graph. To characterize the relationship between the nodes in the graph, an adjacency matrix $ {\rm \mathbf{G}} \in \mathbb{R}^{n \times n} $ is utilized in the PGCN.
			
			According to the spatial proximity based on prior knowledge of neuroscience research, we encode the adjacent nodes in each brain region into a static graph $ {\rm \mathbf{G}_{\mathbf{s}}} \in \mathbb{R} ^ {n\times n} $ with an undirected attribute. Meanwhile, based on the functional connectivity of the brain regions, we extract the spatial information of the EEG signals by multiplying $ {\rm \mathbf{X}} $ left by a trainable weight matrix $ {\rm \mathbf{P}} \in \mathbb{R} ^ {n \times n} $. Then, a bias matrix $ {\rm \mathbf{B}} \in \mathbb{R} ^ {n \times d} $ is added to the above operation to increase the flexibility. To extract the frequency information of the EEG samples, a trainable weight matrix $ {\rm \mathbf{Q}} \in \mathbb{R} ^ {d \times nd} $ is right multiplied with the above, to represent graphs in the $ d $ frequency bands. The overall process of this dynamic graph after reshaping can be expressed as follows:
			
			\begin{equation}
				{\rm \mathbf{G}_{\mathbf{d}}} = Relu[({\rm \mathbf{P}}{\rm \mathbf{X}} + {\rm \mathbf{B}}){\rm \mathbf{Q}}] \in \mathbb{R} ^ {n \times n \times d },
				\label{G^d}
			\end{equation}
			where $ Relu $ is applied to the output to guarantee non-negative elements.
		
		\subsection{Extracting Deep Emotion Features}
		
			After obtaining the dynamic and static graph representations, we constructed a dual-head module to extract the hierarchical characteristics for classification, which enabled the PGCN to progressively learn the discriminative features from coarse- to fine-grained categories. Motivated by the hierarchical characteristic of emotion, a coarse-grained head is proposed to extract the common features of coarse-grained emotion patterns, which has a significant effect on improving the performance of EEG emotion recognition. Meanwhile, a fine-grained head is proposed to extract the information of fine-grained emotions, which is responsible for the target task. Next, we introduce the dual-head module in detail.
			
			To extract high-level emotion features, we employ the graph Fourier transform on the above dual-graph structure based on graph filtering theories~\cite{kipf2017semisupervised}. Owing to the computational complexity of a direct calculation, we adopt Chebyshev polynomials to approximate the graph convolution operation~\cite{kipf2017semisupervised}. Let $ \varphi_{k}(\mathbf{G}) = \mathbf{G}^{k} $ denote the $ k $-order polynomial of the adjacency matrix $ \textbf{G} $. A multilevel graph convolution can be represented as follows:
			
			\begin{equation}
				{\rm \tilde{\mathbf{X}}} = \mathcal{G} * \mathcal{F} = \sum_{k=0}^{K-1}\varphi_{k}({\rm \mathbf{G}}){\rm \mathbf{X}} =  g({\rm \mathbf{G}, \mathbf{X}}) ,
				\label{GCN_op}
			\end{equation}
			where $ \varphi_{k}({\rm \mathbf{G}}) $ is the $ k $-th level graph, and $ {\rm \tilde{\mathbf{X}}} $ is the output. In addition, $ g(\cdot) $ denotes the convolution operation.
			
			In particular, because the PGCN adopts a dual-head module to model the hierarchical characteristic of emotion, we obtain two static ($ {\rm \mathbf{G}^{\mathbf{s}}_{\mathbf{c}}} $ and $ {\rm \mathbf{G}^{\mathbf{s}}_{\mathbf{f}}} $) and two dynamic ($ {\rm \mathbf{G}^{\mathbf{d}}_{\mathbf{c}}} $ and $ {\rm \mathbf{G}^{\mathbf{d}}_{\mathbf{f}}} $) graphs for the coarse- and fine-grained head, respectively. For the coarse-grained head, we modeled the spatial proximity of the EEG data as
			
			\begin{equation}
				{\rm \tilde{\mathbf{X}}^{\mathbf{s}}_{\mathbf{c}}} = g({\rm \mathbf{G}^{\mathbf{s}}_{\mathbf{c}}}, {\rm \mathbf{X}}) \in \mathbb{R} ^ {n \times K^{s}_{c}},
				\label{H(1)S}
			\end{equation}
			where $ K $ is the output dimension for the graph convolution operation, and $ {\rm \mathbf{X}} $ is the input sample. Meanwhile, considering the relationship between emotion and the EEG signals from different frequency bands, the functional connectivity of the brain regions in the coarse-grained head can be captured from multiple graphs, expressed as
			
			\begin{equation}
				\begin{aligned}
					{\rm \tilde{\mathbf{X}}^{\mathbf{d}}_{\mathbf{c}}} &= Cat[ g({\rm \mathbf{G}_{\mathbf{c}}^{\mathbf{d}}}(:,:,1), {\rm \mathbf{X}}),..., g({\rm \mathbf{G}_{\mathbf{c}}^{\mathbf{d}}}(:,:,d), {\rm \mathbf{X}}) ] \\ &~~~~~~~~~~~~~~~~~\in \mathbb{R} ^ {n \times (K^{d}_{c}*d)},
					\label{H(1)D}
				\end{aligned}
			\end{equation}
			where $ Cat[\cdot] $ denotes the concatenation operation. Note that, for the dynamic graph, we consider the different relationships between human emotion and the EEG signals from different frequency bands, leading to differences in the dynamic graph of each frequency band. Similarly, we apply the above operations for the fine-grained head, and obtain the deep features $ {\rm \tilde{\mathbf{X}}^{\mathbf{s}}_{\mathbf{f}}} \in \mathbb{R} ^ {n \times K^{s}_{f}} $ and $ {\rm \tilde{\mathbf{X}}^{\mathbf{d}}_{\mathbf{f}}} \in \mathbb{R} ^ {n \times (K^{d}_{f}*d)} $.
			
			Through the above operation, we can obtain the deep data representations for the coarse- and fine-grained heads, which are denoted as $ {\rm \mathbf{H}_\mathbf{c}} = [{\rm \tilde{\mathbf{X}}^{\mathbf{s}}_{\mathbf{c}}}, {\rm \tilde{\mathbf{X}}^{\mathbf{d}}_{\mathbf{c}}}] \in \mathbb{R}^{n \times (K^{s}_{c} + K^{d}_{c})} $ and $ {\rm \mathbf{H}_\mathbf{f}} = [{\rm \tilde{\mathbf{X}}^{\mathbf{s}}_{\mathbf{f}}}, {\rm \tilde{\mathbf{X}}^{\mathbf{d}}_{\mathbf{f}}}] \in \mathbb{R}^{n \times (K^{s}_{f} + K^{d}_{f})} $, respectively. The overall data representation can then be denoted as $ { \rm \hat{\mathbf{H}}} = [{\rm \mathbf{H}_\mathbf{f}}, {\rm \mathbf{H}_\mathbf{c}}] \in \mathbb{R}^{n \times \hat{K}} $ and $ \hat{K} = K^{d}_{f} + K^{s}_{f} + K^{d}_{c} + K^{s}_{c} $, which progressively learn the discriminative features from coarse- to fine-grained categories. In summary, the proposed dual-head module extracts more discriminative features, enabling the PGCN to achieve advanced results in dual-granular emotion classification.
		
		\subsection{Dual-granular Discriminative Prediction}
		
			As in most supervised models, we add a supervision term to the network by applying the softmax function to the output features, to predict the class label. Specifically, we re-arrange $ {\rm \mathbf{H}_\mathbf{c}} $ and $ { \rm \hat{\mathbf{H}}} $ into one-dimensional vectors $ \mathbf{h_{c}} \in \mathbb{R}^{1 \times [n * (K^{s}_{c} + K^{d}_{c})]} $ and $ \mathbf{\hat{h}} \in \mathbb{R}^{1 \times (n * \hat{K})} $, respectively. The output of the FC layer can be expressed as follows:
			
			\begin{equation}
				{\rm \mathbf{O}_\mathbf{c}} = \mathbf{h_{c}W_{c} + b_{c}} = \{ o_{1}, o_{2}, \dots, o_{P_c}\} \in \mathbb{R}^{1 \times P_{c}},
			\end{equation}
		
			\begin{equation}
				{\rm \mathbf{O}} = \mathbf{\hat{h}W + b} = \{ \hat{o}_{1}, \hat{o}_{2}, \dots, \hat{o}_{P}\} \in \mathbb{R}^{1 \times P},
			\end{equation}
			where $ \mathbf{W_{c}} \in \mathbb{R}^{[n * (K^{d}_{c} + K^{s}_{c})] \times P_{c}} $ and $ \mathbf{b_{c}} \in \mathbb{R}^{1 \times P_{c}} $ are the transform matrices of the coarse-grained head; $ \mathbf{W} \in \mathbb{R}^{(n * \hat{K}) \times P} $ and $ \mathbf{b} \in \mathbb{R}^{1 \times P} $ are the transform matrices of the fine-grained head; $ P_{c} $ and $ P $ represent the numbers of coarse- and fine-grained emotion categories, respectively.
			
			Then, the discriminative prediction from the softmax layer for emotion recognition can be expressed as follows:
			
			\begin{equation}
				\bm{Y_c}(p|\mathbf{X_t}) = exp(o_{p}) / \sum_{i=1}^{P_c}exp(o_{i}),
			\end{equation}
			\begin{equation}
				\bm{Y}(p|\mathbf{X_t}) = exp(\hat{o}_p) / \sum_{i=1}^{P}exp(\hat{o}_i),
			\end{equation}
			where $ \bm{Y_c}(p|\mathbf{X_t}) $ and $ \bm{Y}(p|\mathbf{X_t}) $ denote the predicted probability that the input sample $ \mathbf{X_t} $ belongs to the $ p $th class in the coarse- and fine-grained heads, respectively. Consequently, labels $ l_t $ and $ \hat{l}_t $ of sample $ \mathbf{X_t} $ are predicted as follows:
			\begin{equation}
				l_t = arg \max \limits_p \bm{Y_c}(p|\mathbf{X_t}),
			\end{equation}
		
			\begin{equation}
				\hat{l}_t = arg \max \limits_p \bm{Y}(p|\mathbf{X_t}),
			\end{equation}
			
			Consequently, the loss functions of the dual-granular $ L_{c} $ and $ L $ can be expressed as
			\begin{equation}
				L_{c} = \sum_{t=1}^{M_1}\sum_{p=1}^{P_c}-\tau(l_g, l_t) \times log\bm{Y_c}(p|\mathbf{X_t})
				\label{coarse_loss},
			\end{equation}
		
			\begin{equation}
				L = \sum_{t=1}^{M_1}\sum_{p=1}^P-\tau(\hat{l}_g, \hat{l}_t) \times log\bm{Y}(p|\mathbf{X_t}),
			\end{equation}
			\begin{equation}
				\tau(x,y) =  
				\begin{cases}
					1,& if~x = y \\
					0,&otherwise.
				\end{cases}
			\end{equation}
			Here, $ l_g $ and $ \hat{l}_g $ represent the ground-truth labels of sample $ \mathbf{X_t} $, corresponding to coarse- and fine-grained heads, respectively; $ M_1 $ is the number of training samples.
			
			The PGCN was optimized by iteratively minimizing $ L_{c} $ and $ L $. The procedure used to train the PGCN is presented in Algorithm~\ref{algorithm}. Note that the coarse-grained head only plays an auxiliary role, which is not trained in the same proportion as the fine-grained head. In addition, the coarse-grained head does not affect the weight the update of the fine-grained head. Therefore, the optimization processes of the two heads were isolated from each other.
			
			\begin{algorithm}[h]
				\caption{Procedure of the PGCN model.}
				\label{algorithm}
				\begin{algorithmic}[1]
					\State Initialize model parameters;
					\State Predefine the static graph $ {\rm \mathbf{G}_{\mathbf{s}}} $;
					\For{each $ i \in [0,epochs] $}
						\State Calculate $ {\rm \mathbf{G}_{\mathbf{f}}^{\mathbf{d}}} $ with $ {\rm \mathbf{P}}_{\rm \mathbf{f}}, {\rm \mathbf{Q}}_{\rm \mathbf{f}},{\rm \mathbf{B}}_{\rm \mathbf{f}} $, and $ {\rm \mathbf{X}} $;
						\State Calculate $ {\rm \mathbf{G}_{\mathbf{c}}^{\mathbf{d}}} $ with $ {\rm \mathbf{P}}_{\rm \mathbf{c}}, {\rm \mathbf{Q}}_{\rm \mathbf{c}},{\rm \mathbf{B}}_{\rm \mathbf{c}} $, and $ {\rm \mathbf{X}} $;
						\State Conduct fine-grained dynamic graph convolution and
						\Statex \ \ \ \ static graph convolution; then concatenate them:
						$$ {\rm \mathbf{H}_\mathbf{f}} = [{\rm \tilde{\mathbf{X}}^{\mathbf{s}}_{\mathbf{f}}}, {\rm \tilde{\mathbf{X}}^{\mathbf{d}}_{\mathbf{f}}}]. $$
						\State Conduct coarse-grained dynamic graph convolution
						\Statex \ \ \ \ and static graph convolution; then concatenate them:
						$$ {\rm \mathbf{H}_\mathbf{c}} = [{\rm \tilde{\mathbf{X}}^{\mathbf{s}}_{\mathbf{c}}}, {\rm \tilde{\mathbf{X}}^{\mathbf{d}}_{\mathbf{c}}}]. $$
						\State Make prediction of coarse-grained head from $ \mathbf{H_{c}} $, and
						\Statex \ \ \ \ calculate the loss of coarse-grained head;
						\State Concatenate the features from coarse-grained head and 
						\Statex \ \ \ \ fine-grained head, $$ { \rm \hat{\mathbf{H}}} = [{\rm \mathbf{H}_\mathbf{f}}, {\rm \mathbf{H}_\mathbf{c}}]. $$
						\State Make prediction of fine-grained head from $ \mathbf{\hat{H}} $, and
						\Statex \ \ \ \ calculate the loss of fine-grained head with gradient 
						\Statex \ \ \ \ descent;
						\State Update parameters in the fine-grained head;
						\If{i \% steps == 0 do}
							\State Update parameters in the coarse-grained head
							\Statex \ \ \ \ \ \ \ \ \ with gradient descent;
						\EndIf
					\EndFor
				\end{algorithmic}
			\end{algorithm}
			
	\section{Experiments}
	\label{Sec: Experiment result}
	
		\subsection{Experimental Ssettings}
		\label{Experiment setting}
			\subsubsection{Datasets}
			
				SEED-IV~\cite{8283814} contains 15 subjects in total. These subjects were asked to view 72 movie clips in a comfortable environment. These movie clips easily trigger four emotions: happiness, sadness, fear, and neutrality. Each person watched 24 short movie clips during 3 different sessions. The dataset contains EEG and eye movement data, among which the EEG data are collected using a 62-channel ESI neuroscan system\footnote{https://compumedicsneuroscan.com/}. The locations of the EEG electrodes are based on the international 10–20 system. We divide each session into 1-s segments, and each segment is a training sample. We extract the differential entropy (DE) features at five frequency bands of all training samples.

				The MPED~\cite{song2019mped} collects four modal physiological signals: EEG, galvanic skin response, respiration, and electrocardiogram (ECG). The dataset comprises 30 subjects, each of whom watched 28 videos. The 28 videos were divided into 7 categories: joy, funny, anger, fear, disgust, sadness, and neutral emotion. Herein, we only used the EEG data for emotion recognition. Similarly, we extracted the short-time Fourier transform (STFT) features at five frequency bands from the EEG data.
				
				For the PGCN model, because the coarse-grained head extracts the commonness of the EEG data distribution characteristics from positive, negative, and neutral emotions, we need to relabel each EEG sample by referring to the coarse-grained category. Herein, we consider the coarse-grained correlation of emotions, the characteristic of discrete emotions, and the response of the emotion expression activity in the brain, to relabel the coarse-grained labels according to~\cite{2020CPED, doi:10.1080/02699931.2012.683852, KRAGEL2016444}. Specifically, for SEED-IV, sadness and fear are regarded as negative emotions; happiness is regarded as a positive emotion; and neutral emotion remains unchanged. For the MPED, joy and funny are regarded as positive emotions; anger, fear, disgust, and sadness are regarded as negative emotions; and neutral emotion is unchanged.
			
			\subsubsection{Experiment protocol}
				There are two common experimental protocols in EEG emotion recognition tasks: subject-dependent and subject-independent. For the subject-dependent experiment, we adopted the same experimental protocol as in~\cite{8283814} and~\cite{song2019mped}. Specifically, for SEED-IV, the first 16 trials per session of each subject were used as the training data, and the remaining 8 trials were regarded as the test data. For the MPED, the first 21 trials of EEG data were adopted as the training data, and the remaining 7 trails were used as testing data for each subject. For the subject-independent experiment, we adopted the leave-one-subject-out (LOSO) cross-validation strategy~\cite{DBLP:conf/ijcai/ZhengL16} to evaluate the proposed PGCN model. In the LOSO strategy, the EEG signals of one subject are regarded as testing data, and those of the remaining subjects are allocated as training data. We repeated this procedure for each subject until all EEG signals were treated as the test data once.
				
			\subsubsection{Implementation details}
			
				For the PGCN model, the number of EEG channels $ n $ in the datasets was 62. The number of frequency bands $ d $ was 5. The order of the graph convolution $ K $ was set to 5. The output dimensions of the dynamic and static graph convolution were set to 512 and 256, respectively. In addition, the learning rate was set to $ 10^{-4} $ during the training process. Finally, the entire model was implemented on TensorFlow\footnote{https://tensorflow.google.cn/}.
			
		\subsection{Experiment Results}
			
			\subsubsection{Experimental results for EEG emotion recognition}
		
				To evaluate the classification superiority of the PGCN, we conducted extensive experiments on SEED-IV and the MPED. To compare the PGCN with other representative methods in previous studies on emotion recognition, we also conducted the same experiments using seven other methods: linear SVM~\cite{suykens1999least}, A-LSTM~\cite{song2019mped}, DANN~\cite{ganin2016domain}, DGCNN~\cite{song2018eeg}, IAG~\cite{2020Instance}, BiDANN~\cite{8567966}, and BiHDM~\cite{9105104}. We directly quote (or reproduce) their results from the literature to ensure a convincing comparison with the proposed method. We used mean accuracy (ACC) and standard deviation (STD) as the evaluation criteria for all subjects in the dataset. The experiment results are presented in Table~\ref{Table: results}.
				
				\begin{table}[h]
					\centering
					\caption{Classification performance for EEG emotion recognition on SEED-IV and MPED.}
					\renewcommand{\arraystretch}{1.3}
					\resizebox{\linewidth}{!}
					{		
						\begin{tabular}{|c|c|c|c|c|}
							\hline
							\multirow{3}{*}{\textbf{Method}} & \multicolumn{4}{c|}{\textbf{ACC / STD (\%)}} \\ \cline{2-5}
							
							& \multicolumn{2}{c|}{\textbf{SEED-IV}}          &\multicolumn{2}{c|}{\textbf{MPED}}   \\ \cline{2-5}
							
							& subject-dependent & subject-independent & subject-dependent & subject-independent \\ \hline
							
							SVM~\cite{suykens1999least}& ~56.61/20.05& ~39.39/12.40& ~32.39/09.53&~20.89/03.62\\ \hline
							A-LSTM~\cite{song2019mped}& ~69.50/15.65& ~55.03/09.28& ~38.99/07.53&~24.06/04.58\\ \hline
							DANN~\cite{ganin2016domain}& ~63.07/12.66& ~47.59/10.01& ~35.04/06.52&~22.36/04.37\\ \hline
							DGCNN~\cite{song2018eeg}& ~69.88/16.29& ~52.82/09.23& ~32.37/06.08&~25.12/04.20\\ \hline
							IAG~\cite{2020Instance}& ~73.27/16.15& ~62.64/10.25& ~40.38/08.75&~27.11/05.55\\ \hline
							\textbf{BiDANN*}~\cite{8567966}& ~70.29/12.63& ~65.59/10.39& ~37.71/06.04&~25.86/04.92\\ \hline
							\textbf{BiHDM*}~\cite{9105104}& ~74.35/14.09& ~69.03/08.66& ~40.34/07.59&~{28.27/04.99}\\ \hline
							PGCN &~\textbf{77.08/12.43}&~\textbf{69.44/10.16} &~\textbf{43.56/08.21}&~\textbf{28.39/05.02}   \\ \hline
						\end{tabular}
					}
					\label{Table: results}
				\end{table}
				
				Table~\ref{Table: results} demonstrates that the proposed PGCN model achieves the best performance across all tasks, both subject-dependent and subject-independent, thus verifying the effectiveness of the PGCN. In particular, for the results of the subject-dependent experiments, the proposed method improves the performance of the state-of-art methods BiHDM and IAG by 3\%. Note that the BiHDM and BiDANN use unlabeled test data for training, which improves EEG emotion recognition by narrowing the domain gap between the training and test data. However, in real applications, it is inconvenient to collect test data, even without emotional labels. For the PGCN, it was unnecessary to access any information on the test data during the training process. Despite this, the PGCN achieves better performance than the BiHDM, which further verifies the effectiveness and applicability of our model. Additionally, we can see that the compared IAG method, which also considers the dynamic connection relationships between the EEG electrodes, achieves the overall closest performance to the proposed PGCN. As the main difference between the IAG and PGCN methods, the former considers presenting different graphic representations determined by different input instances, and employs an additional branch to characterize the intrinsic dynamic relationships between different EEG channels. By contrast, the latter (PGCN) focuses on the dynamic functional connectivity of the brain regions from the input instances and the static graphic representation based on the spatial proximity of the EEG electrode distribution, and employs a progressive learning strategy to extract the hierarchical characteristic of emotion. The results of both the IAG and PGCN approaches indicate the importance of considering the dynamic functional connectivity of the brain regions for EEG emotion recognition. Meanwhile, the better performance of the PGCN is attributed to the spatial proximity of the EEG electrode distribution when modeling the EEG emotional signals and exploiting the hierarchical characteristic of emotion when extracting the deep discriminative features.
				
				For subject-independent experiments, the distribution gap is much larger than that in the subject-dependent experiments. In this case, some methods employ a domain adaptation strategy to achieve a more promising performance. The state-of-art BiHDM method adopts the domain adaptation technique to train the model with labeled training data and unlabeled test data. Nevertheless, PGCN still achieves the highest classification results, without using the domain adaptation strategy, and is more suitable for real applications.
				
				Specifically, the SVM only linearly classifies the input features. The graph-based DGCNN and IAG methods consider the dynamic functional connection relationship of the brain regions, but ignore the importance of the physical proximity of the brain in space. The PGCN progressively learns discriminative EEG features, from coarse- to fine-grained features, and then, considers the functional connectivity information and spatial distance information of the brain. Compared with the domain adaptation methods, BiHDM and BiDANN, the PGCN achieves better performance through its reasonable graph representation and model construction, without using unlabeled test data. In summary, the PGCN achieves the best performance for all tasks. These results verify the effectiveness of the dual-head structure and dual-graph construction.
				
			\subsubsection{Confusion matrix based on PGCN results on two datasets}
				
				To explore which emotion is more easily recognized by the proposed model, we depict confusion matrices based on the results of the PGCN on SEED-IV and the MPED, which are shown in Figs.~\ref{3_a} and~\ref{3_b}, respectively. We made the following two observations.
				
				\begin{itemize}
					\begin{figure}[htb]
						\centering 
						\subfigure[Subject-dependent experiment]{\includegraphics[width=0.9\linewidth]{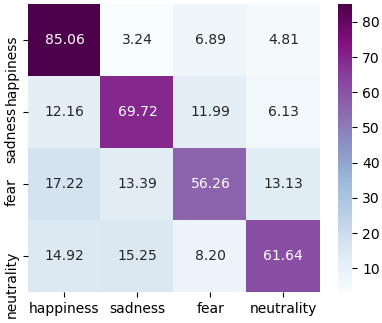}}
						\quad
						\subfigure[Subject-independent experiment]{\includegraphics[width=0.9\linewidth]{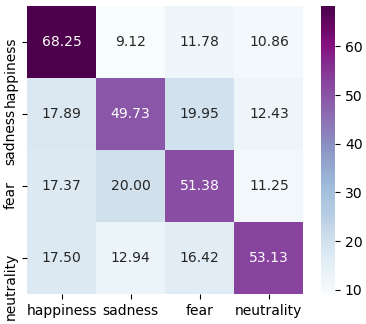}}
						\caption{\label{3_a}Confusion matrix of PGCN results on SEED-IV.}
					\end{figure}
					
					\item[(1)] For SEED-IV in Fig.~\ref{3_a}, happiness is easier to distinguish between subject-dependent and subject-independent experiments, indicating that happiness is more easily induced. Moreover, compared with the results of Fig.~\ref{3_a} (a), we can observe that the recognition rate of fear decreases the least in Fig.~\ref{3_a} (b), indicating that fear is associated with a more similar brain reflection across different people than other emotions. In the subject-independent experiment in Fig.~\ref{3_a} (b), fear and sadness were more likely to be confused. The rate of sadness being mistakenly recognized as fear or fear being mistaken as sadness is as high as 20\%. This indicates that fear and sadness, which are both negative emotions, have a certain degree of commonality, which is consistent with our research perspective.
					
					\begin{figure*}[htb]
						\centering 
						\subfigure[Subject-dependent experiment]{\includegraphics[width=0.48\linewidth]{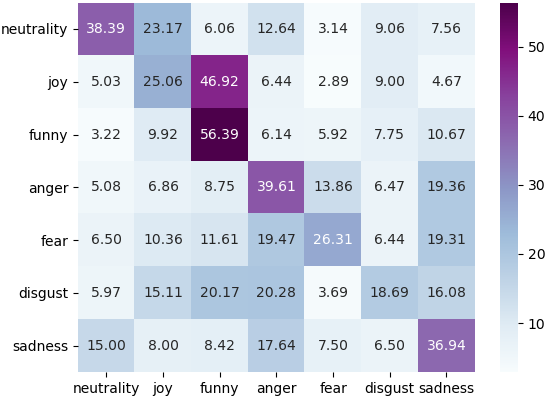}}
						\quad
						\subfigure[Subject-independent experiment]{\includegraphics[width=0.48\linewidth]{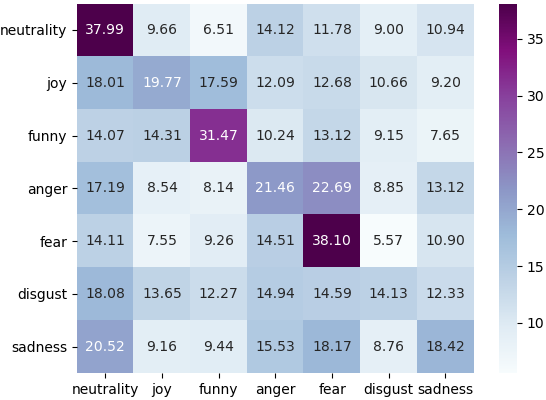}}
						\caption{\label{3_b}Confusion matrix of PGCN results on MPED.}
					\end{figure*}
				
					\item[(2)] For the MPED, there are two types of positive emotions and four types of negative emotions, which is much more complicated than SEED-IV. From the subject-dependent result in Fig.~\ref{3_b} (a), we can see that neutrality, funny, anger, and sadness are easier to distinguish, which has a large gap in the appearance of emotions. Meanwhile, we can observe that joy is more easily confused with funny, which indicates that these two types of emotions have a greater degree of similarity. In addition, funny and happiness may present progressive emotional relationships. When funny is generated, people cannot help but generate joy, making it extremely easy to confuse these two emotions. Among the four negative emotions, anger and fear are more likely to be mistaken for each other. Negative emotions, except anger, will more or less trigger anger in the subjects. The above observations verify the necessity of studying the hierarchical characteristics of emotions. In the subject-independent experiment on the MPED (Fig.~\ref{3_b} (b)), neutrality, funny, and fear with large gaps in the emotional response have a higher prediction reliability, which reflects the small individual differences corresponding to these three types of emotions. The recognition of fear achieves the best performance, which is the same as that observed in SEED-IV. Fear is associated with a more similar brain reflection in different people than other emotions.
				\end{itemize}
		
		\subsection{Discussion}
		\label{Discussion}
			\subsubsection{Effect of dual-head module in PGCN}
			
				To verify the effectiveness of proposed dual-head module, we modify the dual-head module by removing the auxiliary coarse-grained head, which is called PGCN-f. The results for the MPED and SEED-IV are listed in Table~\ref{Table: Ablation study 1 results}.
				
				\begin{table}[htb]
				\centering
				\caption{Experiment results of PGCN-f on two datasets.}
				\renewcommand{\arraystretch}{1.3}
				\resizebox{\linewidth}{!}
				{		
					\begin{tabular}{|c|c|c|c|c|}
						\hline
						\multirow{3}{*}{\textbf{Method}} & \multicolumn{4}{c|}{\textbf{ACC / STD (\%)}} \\ \cline{2-5}
						
						& \multicolumn{2}{c|}{\textbf{SEED-IV}}          &\multicolumn{2}{c|}{\textbf{MPED}}   \\ \cline{2-5}
						
						& subject-dependent & subject-independent & subject-dependent & subject-independent \\ \hline
						PGCN &\textbf{77.08/12.43}&\textbf{69.44/10.16} &\textbf{43.56/08.21}&\textbf{27.88/04.79}   \\ \hline
						PGCN-f &{74.88/14.84}&{60.25/10.89} &{35.98/08.29}&{25.90/04.72}   \\ \hline
					\end{tabular}
				}
				\label{Table: Ablation study 1 results}
				\end{table}
			
				The ACC of the PGCN-f decreases significantly compared with that of the PGCN, demonstrating that the dual-head structure of the PGCN is helpful for improving EEG emotion recognition. Furthermore, we can observe that the performance degradation is more noticeable on the MPED with seven emotions, confirming that the proposed dual-head PGCN model is more suitable for fine-grained classification problems with more apparent hierarchical characteristics. Meanwhile, the PGCN-f achieves a performance similar to that of the state-of-art BiHDM method for subject-dependent experiments on SEED-IV, indicating that the construction of graphs is highly effective in the modeling of EEG signals.
			
			\subsubsection{Effect of graph constructions}
			
				In PGCN, we construct a dual-graph module to model the functional and spatial relationships of the brain regions. To measure the contribution of the different graph constructions, we modify the PGCN by ablating its static and dynamic graph convolution, which are abbreviated as PGCN-d and PGCN-s, respectively. Table~\ref{Table: Ablation study 2 results} shows the subject-dependent and subject-independent experiment results of the two decomposed models on SEED-IV and the MPED.
			
				\begin{table}[htb]
					\centering
					\caption{Classification performance of PGCN-d and PGCN-s.}
					\renewcommand{\arraystretch}{1.3}
					\resizebox{\linewidth}{!}
					{		
						\begin{tabular}{|c|c|c|c|c|}
							\hline
							\multirow{3}{*}{\textbf{Method}} & \multicolumn{4}{c|}{\textbf{ACC / STD (\%)}} \\ \cline{2-5}
							
							& \multicolumn{2}{c|}{\textbf{SEED-IV}}          &\multicolumn{2}{c|}{\textbf{MPED}}   \\ \cline{2-5}
							
							& subject-dependent & subject-independent & subject-dependent & subject-independent \\ \hline
							PGCN &\textbf{77.08/12.43}&\textbf{69.44/10.16} &\textbf{43.56/08.21}&\textbf{27.88/04.79}   \\ \hline
							PGCN-d &{68.51/15.36}&{55.90/12.28} &{34.74/09.26}&{26.05/04.81}   \\ \hline
							PGCN-s &{60.28/03.82}&{48.84/02.80} &{30.88/02.86}&{22.345/00.86}   \\ \hline
						\end{tabular}
					}
					\label{Table: Ablation study 2 results}
				\end{table}
			
				Table~\ref{Table: Ablation study 2 results} shows that the PGCN-d and PGCN-s have different degrees of decline compared with the complete PGCN. This shows the importance of the spatial proximity of the brain and functional connections of the brain regions. The combination of static and dynamic graphs further enhances the PGCN model for EEG emotion recognition. Additionally, the PGCN-d achieves better performance than the PGCN-s, which validates the importance of the functional connectivity-based dynamic graph. We also found that the recognition rate of the PGCN-s decreased more than that of the PGCN-d; however, the STD was much smaller, indicating that the static graph is less sensitive to the differences between individuals.
			
			\subsubsection{Influence of different frequency bands on EEG emotion recognition}
			\label{influence of frequency}
			
				Some studies have proven that there is a correlation between emotion and EEG frequency~\cite{7104132}. To explore the influence of frequency bands on EEG emotion recognition, we conducted additional experiments based on the PGCN with different frequency bands. The classification results are shown in Table~\ref{Table: Ablation study 3 results}.
				
				\begin{table}[htb]
					\centering
					\caption{Results of single-band experiment based on PGCN.}
					\renewcommand{\arraystretch}{1.3}
					\resizebox{\linewidth}{!}
					{		
						\begin{tabular}{|c|c|c|c|c|}
							\hline
							\multirow{3}{*}{\textbf{Frequency bands (Hz)}} & \multicolumn{4}{c|}{\textbf{ACC / STD (\%)}} \\ \cline{2-5}
							
							& \multicolumn{2}{c|}{\textbf{SEED-IV}}          &\multicolumn{2}{c|}{\textbf{MPED}}   \\ \cline{2-5}
							
							& subject-dependent & subject-independent & subject-dependent & subject-independent \\ \hline
							$ \bm{\delta} $ (1\textendash5) &{56.78/12.85}&\textbf{52.54/12.05} &{20.60/04.40}&{19.29/01.31}   \\ \hline
							$ \bm{\theta} $ (4\textendash8) &{55.17/16.91}&{46.27/08.54} &{19.85/02.03}&{19.11/01.32}   \\ \hline
							$ \bm{\alpha} $ (8\textendash14) &{60.18/16.33}&{45.93/10.12} &{22.27/04.84}&{19.82/02.45}   \\ \hline
							$ \bm{\beta} $ (14\textendash30) &\textbf{68.17/16.21}&{48.44/10.10} &{30.65/08.58}&{24.08/04.61}   \\ \hline
							$ \bm{\gamma} $ (30\textendash50) &{62.47/17.67}&{48.21/07.25} &\textbf{36.74/08.88}&\textbf{26.26/05.40}   \\ \hline
						\end{tabular}
					}
					\label{Table: Ablation study 3 results}
				\end{table}

				\begin{figure*}[h]
					\centering 
					\subfigure[subject-1]{\includegraphics[width=0.17\linewidth]{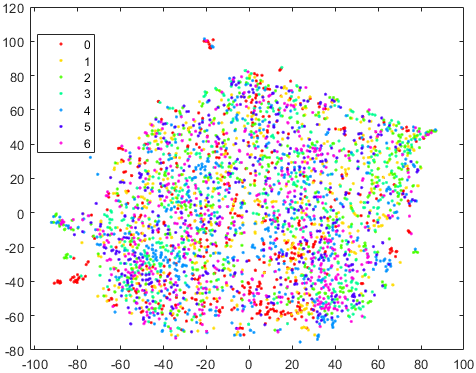}}
					\quad
					\subfigure[subject-2]{\includegraphics[width=0.17\linewidth]{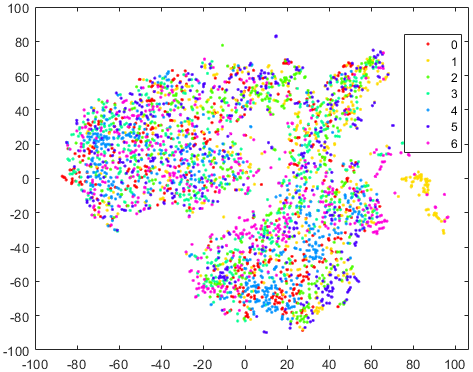}}
					\quad
					\subfigure[subject-3]{\includegraphics[width=0.17\linewidth]{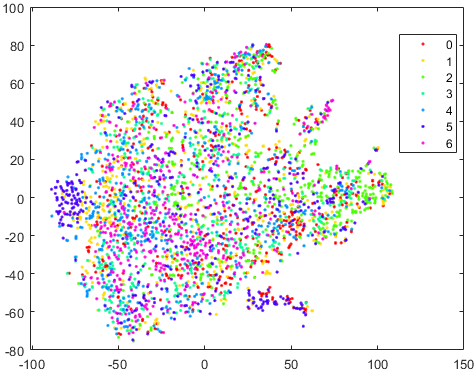}}
					\quad
					\subfigure[subject-4]{\includegraphics[width=0.17\linewidth]{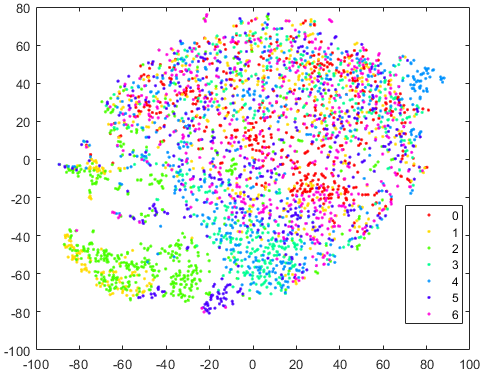}}
					\quad
					\subfigure[subject-5]{\includegraphics[width=0.17\linewidth]{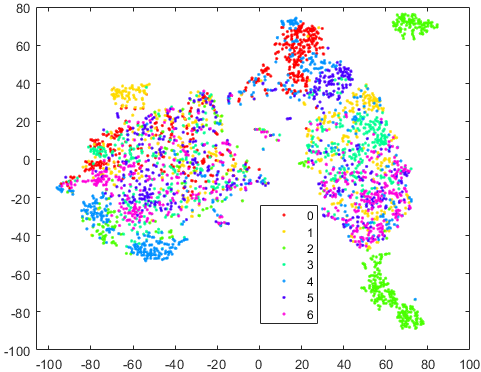}}
					\caption*{(1) Initial state}
					\subfigure[subject-1]{\includegraphics[width=0.17\linewidth]{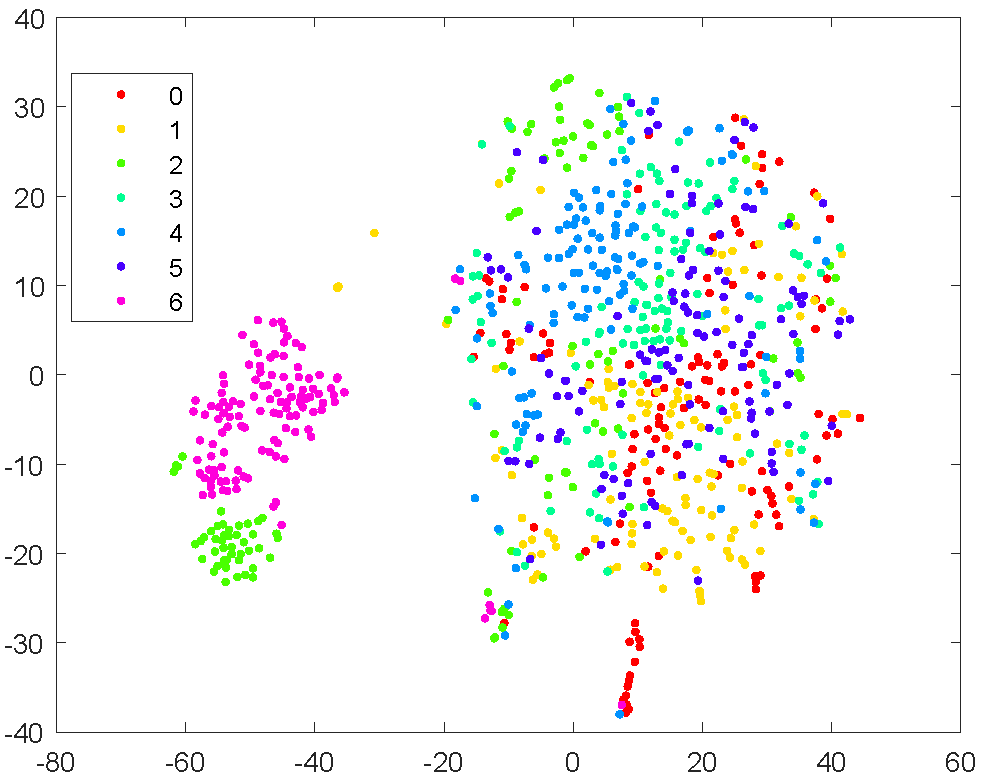}}
					\quad
					\subfigure[subject-2]{\includegraphics[width=0.17\linewidth]{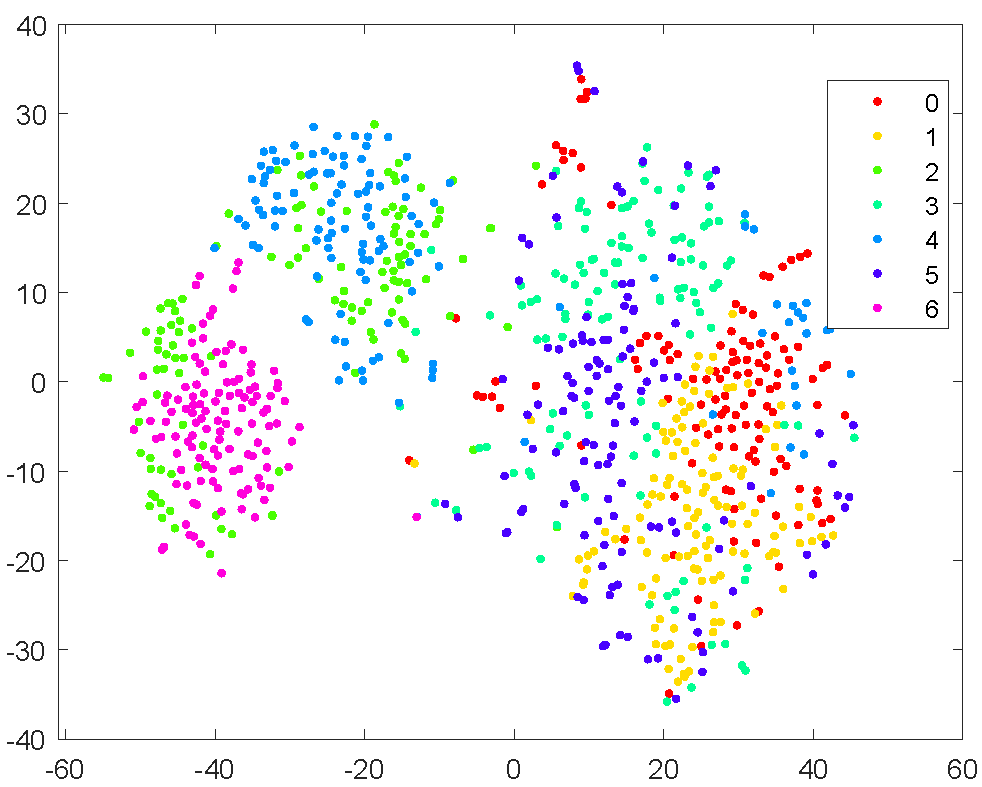}}
					\quad
					\subfigure[subject-3]{\includegraphics[width=0.17\linewidth]{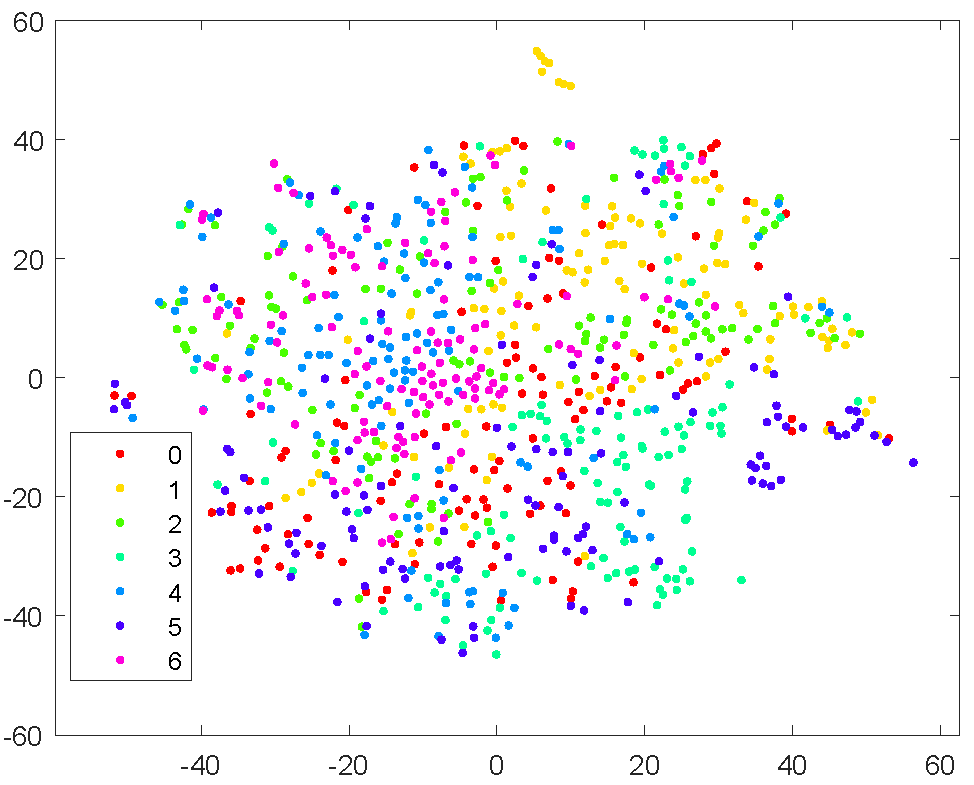}}
					\quad
					\subfigure[subject-4]{\includegraphics[width=0.17\linewidth]{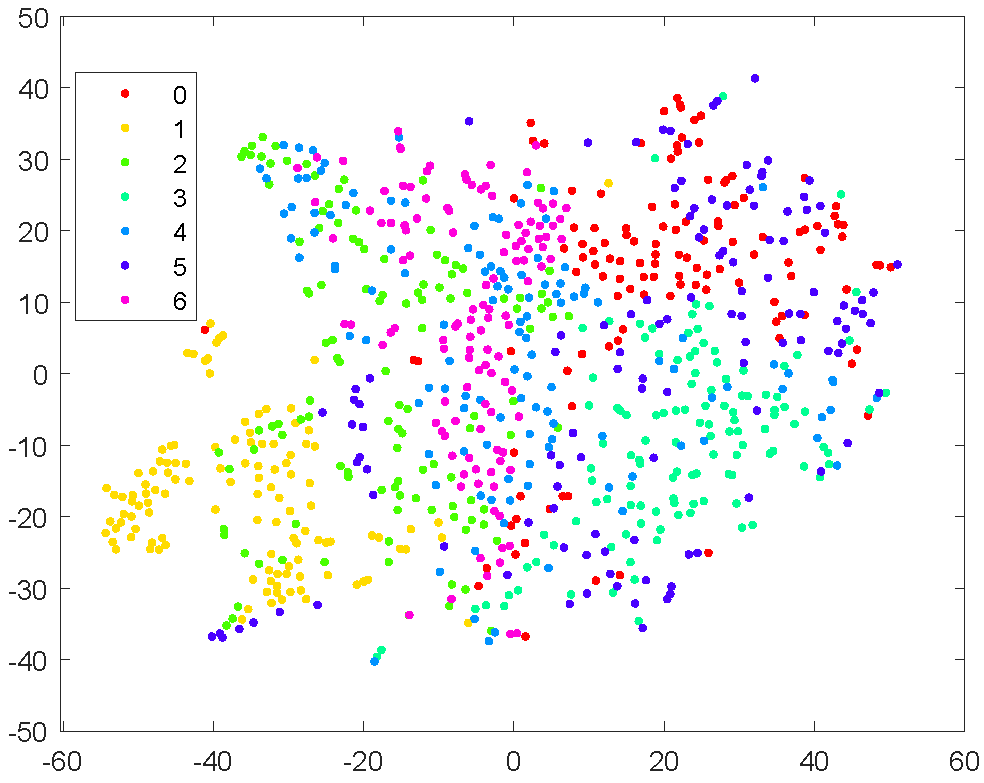}}
					\quad
					\subfigure[subject-5]{\includegraphics[width=0.17\linewidth]{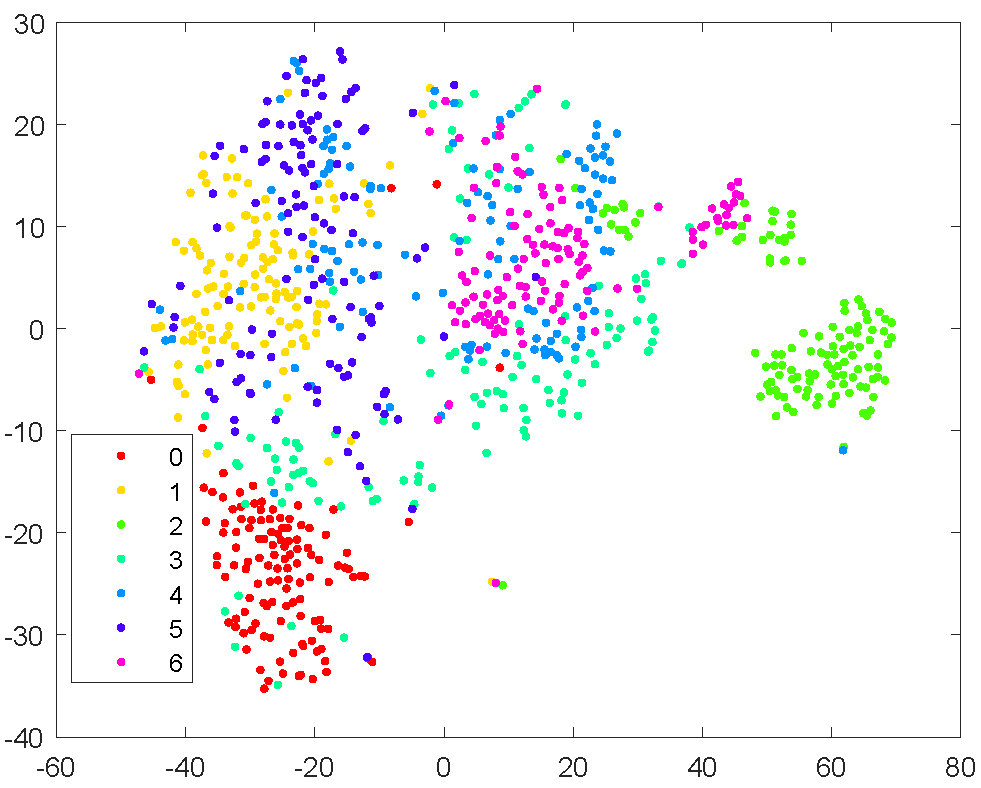}}
					\caption*{(2) Trained state}
					\caption{\label{4}T-SNE visualization of features from PGCN in subject-dependent experiment on MPED. The points marked as 0$ \sim $6 denote neutrality, joy, funny, anger, fear, disgust, and sadness, respectively. (1) Initial state and (2) trained state.}
				\end{figure*}

				The experiment results indicate that the classification ACC of high-frequency EEG data, e.g., for the $ \bm{\beta} $ and $ \bm{\gamma} $ bands, is higher. This result is consistent with the conclusion in~\cite{7104132}, thereby confirming that changes in human emotion are related to the high-frequency band. We can see that the recognition rate of the $ \bm{\delta} $ band in the subject-independent experiment on SEED-IV was the highest among the five bands considered. We speculate that the human brain contains more basic information within the low-frequency band; it is necessary in a subject-independent experiment to explore the commonalities between individuals.
				
			\subsubsection{T-SNE visualization of PGCN}
			
				To further exploit the training process of the PGCN, we employ T-SNE technology~\cite{van2008visualizing} to visualize the training process of the proposed model. As shown in Fig.~\ref{4}, we visualized the EEG data of the first five subjects, from the initial to trained state, based on the MPED. Fig.~\ref{4} (1) shows that the seven types of emotions were mixed and difficult to distinguish during the initial state. After training, the same types of emotions gather, as shown in Fig.~\ref{4} (2). We found that the points representing the emotional EEG signals are easier to distinguish visually than before, which validates the effectiveness of our model.
		
			\subsubsection{Functional connections for emotion expression in the PGCN}
			
\begin{figure*}[h]
				\centering
				\subfigure[ $\delta$ band]{
					\includegraphics[width=0.16\linewidth]{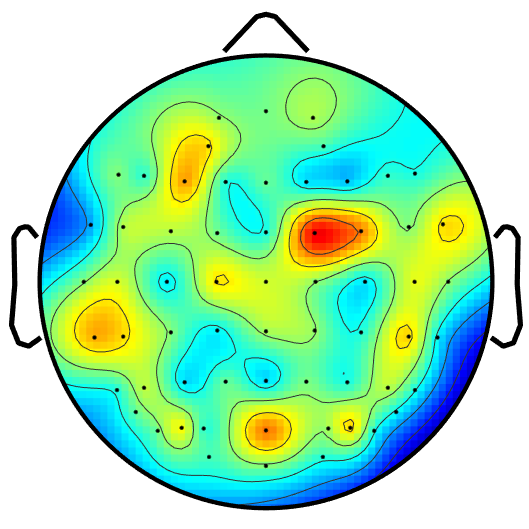}
				}
				\quad
				\subfigure[$\theta$ band]{
					\includegraphics[width=0.16\linewidth]{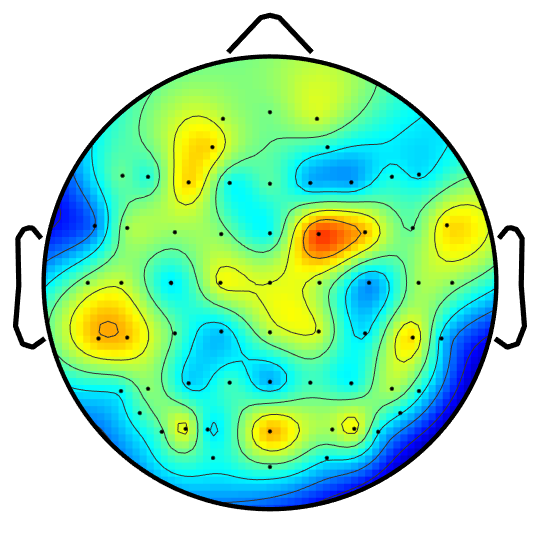}
				}
				\quad
				\subfigure[$\alpha$ band]{
					\includegraphics[width=0.16\linewidth]{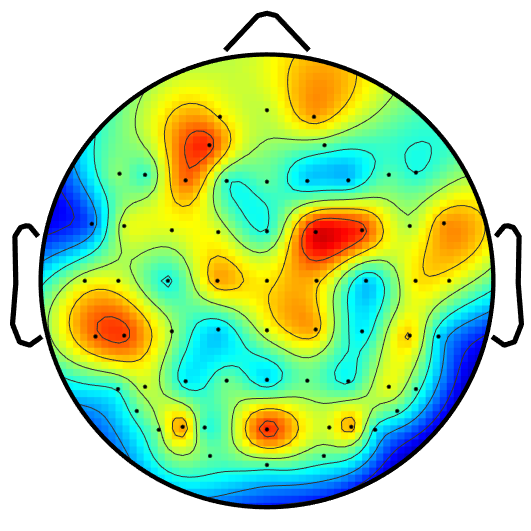}
					%\caption{fig1}
				}
				\quad
				\subfigure[$\beta$ band]{
					\includegraphics[width=0.16\linewidth]{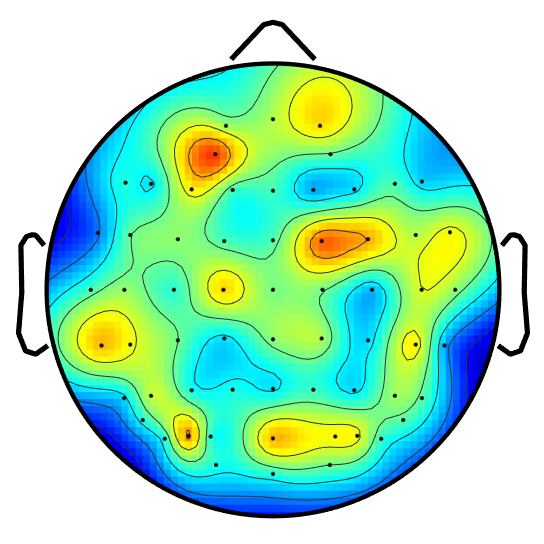}
				}
				\quad
				\subfigure[$\gamma$ band]{
					\includegraphics[width=0.16\linewidth]{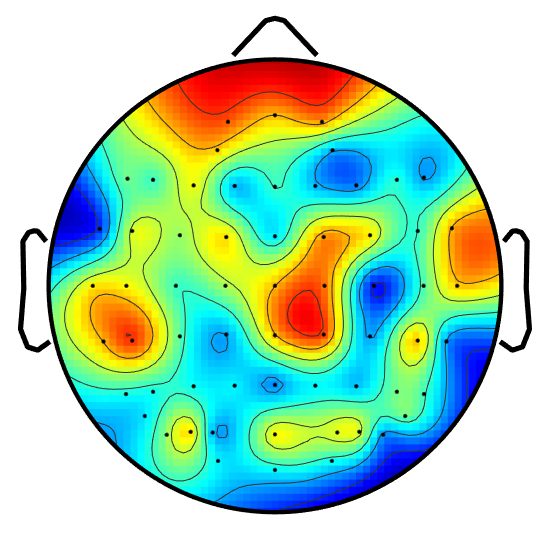}
				}
				\caption{\label{7}Visualization of $ \mathbf{G_{f}^{d}} $ in subject-dependent experiment based on SEED-IV: (a) 1-4, (b) 4-8, (c) 8-14, (d) 14-30, (e) 30-50 Hz. Darker red denotes a more significant contribution.}
			\end{figure*}

				To clearly illustrate the contribution of the dynamic graph in extracting the functional connections of the brain regions, we depict the electrode activity maps from five frequency bands in Fig.~\ref{7}. Their contribution is evaluated by visualizing the fine-grained dynamic graph $ \mathbf{G_{f}^{d}} $, which is predicted from the input samples and distributed on the scalp in the form of electrode nodes. As shown in Fig.~\ref{7}, the distributions of the dynamic graphs within different frequency bands are extremely different. The more significant part concentrates on the high-frequency bands, which is confirmed by the comparative experiment discussed in Section~\ref{influence of frequency}. Moreover, Fig.~\ref{7} (e) shows that EEG data from the frontal and temporal lobes provide more significant contributions. This observation is consistent with the cognition observation in biological psychology~\cite{COAN20047, davis2001amygdala}; namely, the frontal and temporal lobes are the main areas of the brain that are correlated with emotional activity~\cite{7946165}. Moreover, a part of the parietal lobe plays an important role in emotion recognition, which may arise from thalamus activity~\cite{BRITTON2006397}. These observations further confirm that the PGCN concentrates more on the brain regions that have a high correlation with emotions, which can help us explore the relationships between the functional connectivity of the brain regions and different emotions.
				
	\section{Conclusion}
	\label{Sec: Conclusion}
	
		In this study, we propose a PGCN to capture the hierarchical characteristics of EEG emotional signals and progressively learn discriminative EEG features. The designed dual graphs used in the PGCN characterize the intrinsic relationships between different EEG channels, which contain the dynamic functional connections and static spatial proximity information of brain regions from neuroscience research. The followed dual-head module enables the PGCN to progressively learn more discriminative EEG features, from coarse- (easy) to fine-grained categories (difficult), by referring to the hierarchical characteristic of the emotion. Extensive experiments on two public EEG emotion datasets demonstrate that the proposed PGCN method achieves state-of-art performance. The better recognition performance of the PGCN is attributed to the fact that it can model the EEG signals based on the dynamic functional connections and static spatial proximity of the brain regions, while using the hierarchical characteristic of emotion to progressively capture the discriminative features. Using the PGCN, we also investigate the contribution of the dual-head module and dual-graph for EEG emotion recognition, and explore the relationships between frequency bands and emotion. In a future study, we will further investigate the hierarchical information of emotion and the dynamic functional connectivity of the brain regions.
	
	\bibliographystyle{IEEEtran}
	\newpage
	\bibliography{ref_bib}
\end{document}